\documentclass[runningheads]{llncs}
\usepackage{mathtools,etoolbox}
\usepackage{graphicx}
\usepackage{hhline}
\usepackage{multirow}
\usepackage{booktabs}
\usepackage{caption}
\usepackage{subcaption}
\usepackage{svg}
\usepackage{amsmath,amsfonts,bm} % Math packages
\begin{document}
%
%\title{How Clinically-relevant Secondary Features Lead to Improved Pancreatic Tumor Detection}
\title{Improved Pancreatic Tumor Detection by Utilizing Clinically-Relevant Secondary Features}
%
%\titlerunning{Abbreviated paper title}
% If the paper title is too long for the running head, you can set
% an abbreviated paper title here
%
\author{Christiaan G.A. Viviers\inst{1} \and Mark Ramaekers\inst{2} \and Peter H.N. de With\inst{1}, Dimitrios Mavroeidis\inst{3}
\and Joost Nederend\inst{2}, Misha Luyer\inst{2} \and Fons van der Sommen\inst{1}}
%
%index{Viviers, Christiaan}
%index{Ramaekers, Mark}
%index{de With, Peter}
%index{Mavroeidis, Dimitrios}
%index{Nederend, Joost}
%index{Luyer, Misha}
%index{van der Sommen, Fons}
%
\authorrunning{C.G.A.Viviers et al.}
% First names are abbreviated in the running head.
% If there are more than two authors, 'et al.' is used.
%
\institute{Eindhoven University of Technology, Eindhoven 5612 AZ, The Netherlands \and Catharina Ziekenhuis, Eindhoven EJ 5623, The Netherlands \and Philips Research,  Eindhoven AE 5656, The Netherlands \newline
\email{c.g.a.viviers@.tue.nl}}

\maketitle              % typeset the header of the contribution
\begin{abstract}
Pancreatic cancer is one of the global leading causes of cancer-related deaths. Despite the success of Deep Learning in computer-aided diagnosis and detection (CAD) methods, little attention has been paid to the detection of Pancreatic Cancer. We propose a method for detecting pancreatic tumor that utilizes clinically-relevant features in the surrounding anatomical structures, thereby better aiming to exploit the radiologist's knowledge compared to other, conventional deep learning approaches. To this end, we collect a new dataset consisting of 99~cases with pancreatic ductal adenocarcinoma (PDAC) and 97~control cases without any pancreatic tumor. Due to the growth pattern of pancreatic cancer, the tumor may not be always visible as a hypodense lesion, therefore experts refer to the visibility of secondary external features that may indicate the presence of the tumor. We propose a method based on a U-Net-like Deep CNN that exploits the following external secondary features: the pancreatic duct, common bile duct and the pancreas, along with a processed CT scan. Using these features, the model segments the pancreatic tumor if it is present. This segmentation for classification and localization approach achieves a performance of 99\% sensitivity (one case missed) and 99\% specificity, which realizes a 5\% increase in sensitivity over the previous state-of-the-art method. The model additionally provides location information with reasonable accuracy and a shorter inference time compared to previous PDAC detection methods. These results offer a significant performance improvement and highlight the importance of incorporating the knowledge of the clinical expert when developing novel CAD methods.

\keywords{Pancreatic Cancer \and Tumor Segmentation  \and CNN}
\end{abstract}

\section{Introduction}
Pancreatic cancer is one of the leading causes of cancer-related deaths worldwide with a dismal prognosis and an overall 5-year survival rate of 9\%~\cite{Rahib2014-dh}. Due to late recognition, most patients advance to late stages of the disease or even metastases. Pancreatic tumor detection using CT imaging is considered to be the gold standard for the detection of pancreatic cancer~\cite{Treadwell2016-jr}. The obtained accuracies of pancreatic ductal adenocarcinoma (PDAC) detection using CT imaging or other radiological imaging techniques largely depends on radiological expertise. Lack of such expertise may result in delayed recognition, which is problematic since only 20\% of patients at the time of diagnosis are eligible for resection~\cite{Hidalgo2010-sp}. Therefore, early detection of pancreatic cancer holds significant promise by enabling surgical treatment and improving treatment outcomes.

Initial diagnosis of pancreatic tumors through CT imaging maintains acceptable sensitivity measures of around 90\% for pancreatic cancer diagnosis~\cite{Lee2014-oj}. In general, pancreatic tumors appear hypodense compared to normal pancreatic parenchyma. However, indeterminate CT findings such as small tumor size, growth pattern, iso-attenuating pancreatic cancer and the difficulty in differentiating from chronic pancreatitis, can make accurate delineation of viable tumor tissue a troublesome task~\cite{Ahn2009-hc}. In addition, pancreatic cancer often causes non-specific symptoms prior to developing into an advanced stage. Therefore, it is important to identify secondary features which might indicate disease to improve early detection of PDAC. Computer-aided diagnosis and computer-aided detection (CAD) techniques hold great promise in enabling the early detection of PDAC. Such a tool allows for expert knowledge to be captured and shared, which can be used when the patient is first screened for the disease. Deep learning-based CAD methods have achieved impressive results in recent years. For these methods to be successfully adopted in the clinical environment, it is necessary to provide more than the standard ``black-box" machine learning model~\cite{Kriegsmann2021,PETCH2022204}. For clinical acceptance of this technology, on top of high detection accuracies, it is essential to provide additional insights into the model's operation.

In this research, we propose a PDAC segmentation model that utilizes the same visual cues in the surrounding anatomy that experts use when looking for the presence of PDAC. This focus and way of working is to maximally leverage easily accessible external information and fully exploit clinical expertise, to ultimately optimize classification and localization performance. Since we start from the radiologists' reasoning, our method becomes clinically meaningful. For instance, a clinician pays close attention to pancreatic ductal size as a large (potentially dilated) duct could be indicative of tumor. Compared to normal pancreatic tissue in a CT scan, pancreatic cancer appears less visible as an ill-defined mass. It enhances poorly and is hypodense between 75\% and 90\% of arterial phase CT cases. For this reason, experts utilize secondary features which may be predictive of pancreatic cancer. These include, but are not limited to: ductal dilatation, hypo-attenuation, ductal interuption, distal pancreatic atrophy, pancreatic contour anomalies and common bile duct dilation. For a detailed description of these indicators, we refer to the work by Zhang~\textit{et al.}~\cite{Zhang2018}.

As these secondary features offer crucial information to experts during analysis, we hypothesize that a Deep Learning-based CAD method could also explicitly leverage this information. As such, we enrich the input of a 3D U-Net~\cite{3dunet} segmentation model with an indication of the external secondary features and observe state-of-the-art results in PDAC detection.  
\section{Related work on PDAC detection}
Deep learning has rapidly advanced the development of CAD methods across various domains. Invaluable research towards automated PDAC detection has also been conducted. Recently, both Liu~\textit{et al.}~\cite{LIU2020e303} and Si~\textit{et al.}~\cite{Si2021-uq} implemented a patch-based PDAC classification of CT volumes. These classification methods show high accuracy, but clinicians require more interpretable results, such as an indication to the tumor area. This will enable further insights about the tumor for potential follow-up tasks. To this end, a semantic segmentation approach is more suitable, since it additionally provides tumor localization. A Multi-Scale Coarse-to-Fine Segmentation method is proposed by Zhu~\textit{et al.}~\cite{zhumulti} that makes use of three U-Net-like segmentation models at different resolutions in a segmentation-for-classification approach. The output of the three networks are combined using a connected component graph between the adjacent tumor-positive voxels. Finally, false positive components from the graph are pruned and the tumor voxels are selected based on empirically selected thresholds. We refer to the work by Zhu~\textit{et al.} as previous state-of-the-art with a sensitivity of 94.1\% and a specificity of 98.5\%. Similarly, Alves~\textit{et al.}~\cite{automatic_pdac} proposed a similar segmentation-for-classification approach that makes use of four nnU-Net-based~\cite{nnUnet} models to ultimately detect the presence of the tumor. Although these methods achieve impressive results, it is engineering solutions that lacks transparency, sufficient motivation from a clinical perspective, in addition to suffering from long inference times. We propose a more intuitive, clinically-motivated method for PDAC detection. The proposed approach utilizes clinically-relevant cues to realize state-of-the-art detection scores while significantly simplify the network architecture, making it more suitable for deployment in medical centres.

\section{Methods}
\subsection{Data Collection}
In our retrospective single-center research study, we collected contrast-enhanced CT images of 97~control cases and 99~cases with pancreatic ductal adenocarcinoma (PDAC) located in the pancreatic head from the Catharina Hospital Eindhoven. Patients aged 18~years or above who underwent surgical treatment at the Catharina Hospital Eindhoven for pancreatic head cancer, were eligible when both a surgical report and a complete pathology report were available. All CT-scans were manually annotated in preparation for being used in this research. Relevant anatomical structures (tumor, pancreas, pancreatic duct and common bile duct) were annotated by a surgical resident and supervised by an expert abdominal radiologist using IntelliSpace Portal (Software package available from Philips Healthcare, The Netherlands). Patients in the control group were derived from a previous randomized control trial in which patients with esophageal cancer were included. These patients all had a CT scan as a preoperative work-up.
\begin{figure}[t]
\begin{minipage}{\linewidth}
  \centering  \centerline{\includegraphics[width=1.0\linewidth]{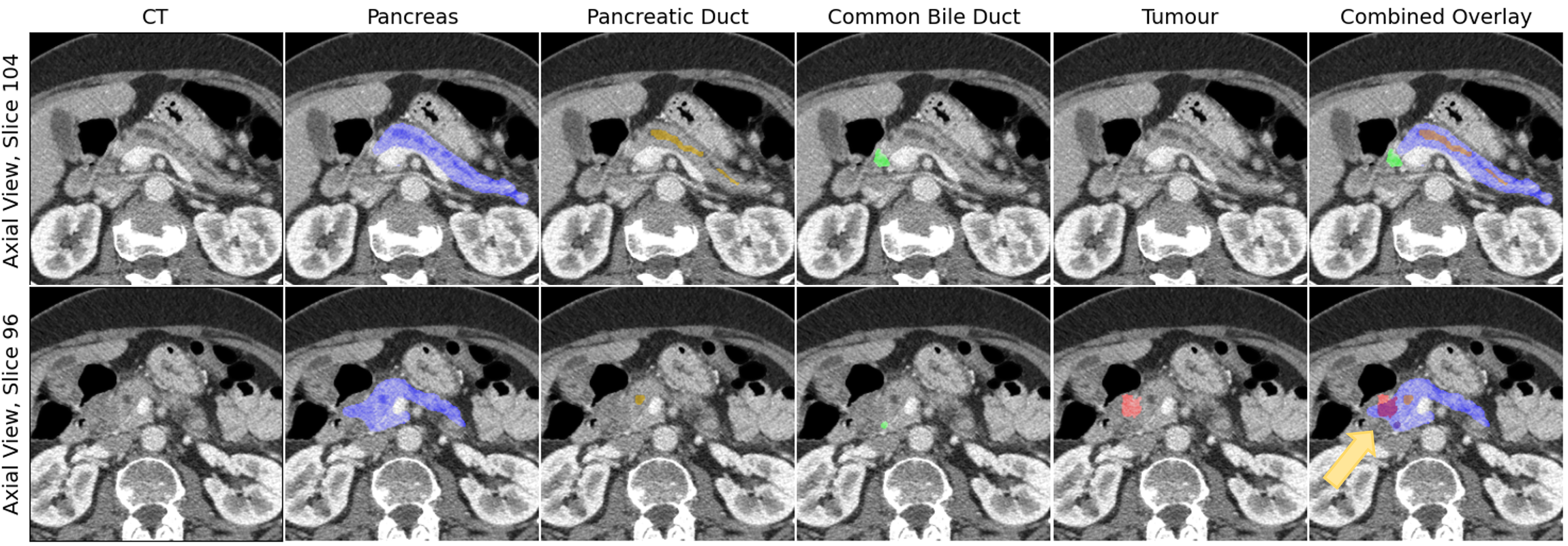}}
  \caption{\textit{Two slices from a case showcasing the involvement of the different structures and dilated ducts caused by the tumor blockage. These features are indicative of the tumor presence and its location. The bottom right image shows the involvement between the tumor and pancreas in dark red.}}
  \label{fig:segmentation}
\end{minipage}
\end{figure}
The external secondary features play an important role in the expert radiologist's decision-making w.r.t. tumor presence, size and location. As such, much annotation effort was spent on not only the tumor, but also these indicative features. Two important factors that arose during this process were (1)~how to annotate some of structures that belong to the same organ (pancreatic duct inside the pancreas) and (2)~defining cases where there is a gradual transition from one structure to the other. The latter occurs when the common bile duct enters the pancreas, but importantly, also with the tumor itself. We decided that each structure should be annotated and stored separately to attain maximum information. However, this implied that CT voxels could potentially belong to multiple structures simultaneously. Figure~\ref{fig:segmentation} depicts an example case and corresponding ground-truth annotations. The last image in the bottom row shows the overlap between the pancreas and the tumor and the pancreatic duct in the pancreas.

As part of the Medical Decathlon~\cite{Medical_Decathlon} (MD), Task 07 involves the segmentation of the pancreas and pancreatic masses (intraductal papillary neoplasms, pancreatic neuroendocrine tumors, or pancreatic ductal adenocarcinoma). This dataset consists of patients with often well developed late-stage disease. As a result, there is a high proportion of large tumors and easily detectable cysts in this dataset. In addition, due to the extensive disease and associated symptoms, many cases contain metal stents, which could induce a bias in a learning algorithm. To the best of our knowledge, this is the only publicly available dataset that aims to detect pancreatic cancer, and although very valuable, it is still a step away from being an ideal dataset for training a deep learning-based CAD system for detection of PDAC. To provide some insight into how our approach competes against other methods on this public benchmark, we have supplemented 10\% of this dataset's training set (28 cases) with suspected adenocarcinoma in the pancreatic head with separate annotations for the pancreatic duct, common bile duct, the full pancreas (unobstructed by the tumor) and the tumour. This subset will be used as an unseen test set in our experiments\footnote{Newly annotated data: https://github.com/cviviers/3D\_UNetSecondaryFeatures}. 
\subsection{Segmentation model for Classification and Localization}
\begin{figure}[t]
\centering
{\resizebox{0.9\linewidth}{!}{
\begin{minipage}{\linewidth}
  \centering  \centerline{\includegraphics[width=1.0\linewidth]{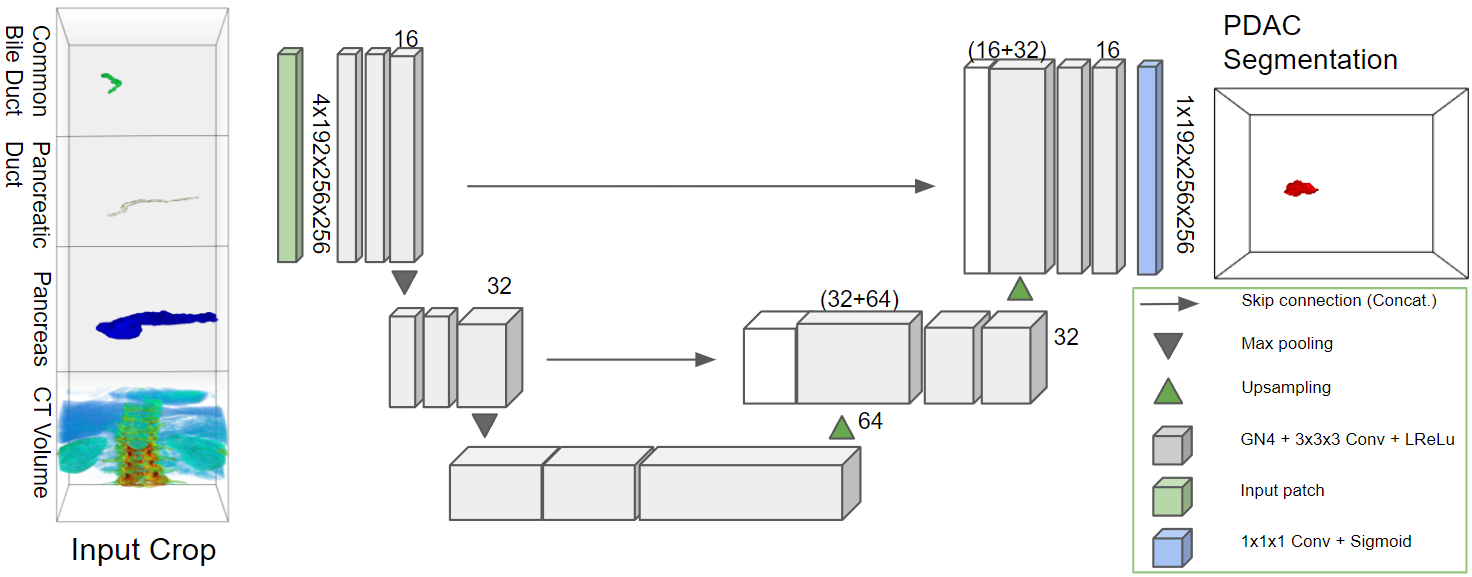}}
  \caption{\textit{Diagram of the 3D U-Net used for tumor segmentation in abdominal CT scans, provided with detailed external secondary feature segmentation maps.}}
  \label{fig:arch}
\end{minipage}}}
\end{figure}

To provide clinicians with the necessary assistance in the early diagnosis of pancreatic cancer, an artificial intelligence system that is capable of accurately detecting PDAC in a clinically-interpretable way needs to be deployed. The system does a first screening, filtering out a large number of normal cases and preserving the cases with an indication of cancer. A CAD system that uses the same indicative features as the clinician could potentially be more explainable as it will provide additional insights as to how the result was derived. This could inspire more trust in the CAD system by the clinician, since it uses the same clinical way of working. As a result, the tumor is more likely to be detected earlier, leading to a better outcome. Bearing this in mind, we develop a segmentation for classification and localization method. We use a standard 3D U-Net (Figure~\ref{fig:arch}) that takes the segmentation maps capturing the external indicators as input and segments the tumor in the CT volume. These external secondary features are used by expert radiologists to identify and localize the tumor, but are also much more easily obtainable and can be identified in a non-expert setting with minimal effort. In practice, these can also be obtained by a prior segmentation model to streamline the process even further. Taking the difficulties related to accurate segmentation of pancreatic tumor into account (even for an expert radiologist), our objective is not to acquire a detailed segmentation map. Instead, we aim for an indication of where the tumor could be located. A detailed segmentation network or radiologist can then initiate follow-up work. 
% investigation or forward the patient and corresponding CT scans to an expert centre.
%  such as forwarding the patient to an expert centre for the necessary treatment
\subsection{Experiments}
To test our approach and the extent of the influence of the external secondary features in the tumor detection, the following experiments are conducted. (1)~We start by setting the baseline at detecting a tumor using only the CT scan. This baseline is set using the popular nnU-Net~\cite{nnUnet} (Full-Resolution 3D) and a custom 3D U-Net. (2)~In a follow-up experiment, we add the detailed segmentation maps of the pancreas and ducts to the CT scan, concatenated channel-wise. The same custom 3D U-Net is trained to segment the tumor, but now with this additional information derived from the radiologist. (3)~As an ablation experiment, we replace the segmentation maps of the ducts with a Boolean input. The pancreatic and common bile ductal 3D volumes are replaced with unity values if they are dilated or not. (4) Finally, we apply the models (using the CT scan and detailed segmentation maps), trained and validated on the three data folds of our dataset, to the Medical Decathlon Dataset as test set.
\subsection{Data Preparation \& Training Details}\label{Sec:DataPrep}
The radiologist starts the investigation for a tumor by localizing the pancreas in the CT scan. Once the pancreas have been located, the radiologist slides through scans looking for the various aforementioned indicative secondary features of the cancer. As such, we preprocess our data according to this expert's way of working. We derive a detailed segmentation map of the pancreas, pancreatic duct and common bile duct from the abdominal CT scan and use it as the secondary features. In practice, this is performed by a prior segmentation model, but since this is outside the scope of this paper, we use the ground-truth detailed segmentations provided by the expert radiologist. We crop the CT scan and corresponding labels, centered around the pancreas' center of mass. The crop is shaped within the dimensions [192, 256, 256] in the $z,x,y$-axes, respectively. Additional resampling and normalization is performed as described in the work by Isensee~\textit{et al.}~\cite{nnUnet} prior to cropping. We stack the CT scan, pancreas and two ducts channel-wise along a 4th dimension in preparation for training. Our final dataset is: $\mathbf{D} = \{(\mathbf{X}_1, \mathbf{Y}_1), ...(\mathbf{X}_N, \mathbf{Y}_N)\}$, with $N$ being the dataset size, $\mathbf{X}_n \in \mathbb{R}^{C \times Z \times W \times H}$ is the 4D volume of input data and $\mathbf{Y}_n \in \mathbb{R}^{ Z  \times W  \times H}$ is the 3D tumor segmentation map.

In our implementation, we perform threefold cross-validation using a random 70/30\% training and validation split and report results on the validation sets and the MD dataset as test set. The custom 3D U-Net is implemented in PyTorch and extends on the work by Wolny~\textit{et~al.}~\cite{eLife}. During training we only employ a cross-entropy loss, a batch size of 2, an Adam optimizer with an initial learning rate of $1\cdot10^{-4}$ and a weight decay of $1\cdot10^{-5}$. We use extensive data augmentation, consisting of random flipping, random rotation, elastic deformation, contrast adjustment, and additive Gaussian and Poisson Noise. In all our experiments, the same crops, hyperparameters and augmentation techniques are used, with a hardware configuration based on a TITAN RTX GPU\footnote{Commercially available from Nvidia Corp., CA, USA}.
\section{Results \& Discussion}
The experimental results are listed in Table~\ref{tab:results}. In all cases, the model outputs are binarized (standard threshold setting of 0.5) and converted to segmentation maps. For the classification metric, if the resulting segmentation prediction overlaps with the ground-truth tumor label, even partially, we consider it a true positive prediction. If there is a yes (no) prediction and a tumor without overlap, it is a false positive (negative). In the case there is no prediction whatsoever and the tumor is absent, we consider it a true negative. The sensitivity, specificity and average Dice (across all the tumor-positive cases) on the validation sets are reported. We also show the results of the model using the full input (CT scan and detailed segmentation maps) and applied to the test MD dataset.
\begin{table}[b]
\centering
{\resizebox{0.9\linewidth}{!}{
\begin{tabular}{l|c|ccc} 
\toprule
 \textbf{Data Input } \enspace & \textbf{Model} & \textbf{Sensitivity}  & \textbf{Specificity} & \textbf{Dice}  \\ 
\hline
CT Only \enspace  &  \enspace  nnU-Net  \enspace & \enspace$ 0.92 \pm 0.02$ \enspace & $0.27 \pm 0.16$\enspace & $0.42 \pm 0.04 $\enspace\\
  CT Only  \enspace  & \enspace3D U-Net  \enspace & \enspace$0.98 \pm 0.03$  \enspace & $0.11 \pm 0.10$ \enspace  &\enspace $0.40\pm 0.07$ \enspace\\
  Binary Ducts  \enspace  & \enspace3D U-Net  \enspace &\enspace $0.83 \pm 0.24$ \enspace & $0.19 \pm 0.06$ \enspace  & \enspace$0.16 \pm 0.04$  \enspace\\
  Full \enspace  & \enspace 3D U-Net    \enspace &\enspace $1.00 \pm 0.00$ \enspace & $0.99 \pm 0.02$ \enspace  & $ 0.31 \pm 0.07$\enspace\\

\bottomrule
Test MD - Full \enspace  & \enspace 3D U-Net    \enspace &\enspace $0.99 \pm 0.02$ \enspace & N/A  & $ 0.31 \pm 0.05$\enspace\\
\bottomrule
\end{tabular}}
\caption{\textit{Results obtained with the nnU-Net and 3D U-Net with different input channel information. Given the limited amount of data, numbers are constrained to two decimals}. \label{tab:results} }}
\end{table}

\emph{Results on Baseline:} The nnU-Net and our 3D U-Net showcase similar performance when trained using only the CT scan as training data. In both models, the network eagerly tries to segment the tumor, even when the tumor is absent, resulting in a low specificity.

\emph{Adding Binary Ducts:} The segmentation performance does not improve when the model is trained using the additional \emph{binary} labels, indicating the presence of dilated ducts. Hence, duct dilation alone is not a tumor-deciding factor and has to be combined with indicative features from the hypodense tumor region.

\emph{Detailed Segmentation Maps:} The model is provided with the detailed segmentation maps of the ducts and pancreas, along with the CT scan. We observe that the model can learn the connection between these indicative features and the presence of a tumor. The model correctly predicts tumor with an overlap of the ground-truth segmentation in the majority of cases. In a single case, tumor is predicted to be at a different location than the label (False Positive). We observe a lower Dice score compared to the models with only CT scans as input. These baseline models maximally predict tumors in most cases. This results in a higher Dice score when there is a tumor factually present, at the expense of a large number of false positive predictions. When increasing sensitivity, the model logically locates more tumors, albeit some with low Dice scores.

\begin{figure}[t]
\centering
{\resizebox{0.9\linewidth}{!}{
\begin{subfigure}{.24\textwidth}
    \centering
    \includegraphics[angle=180, width=.95\linewidth]{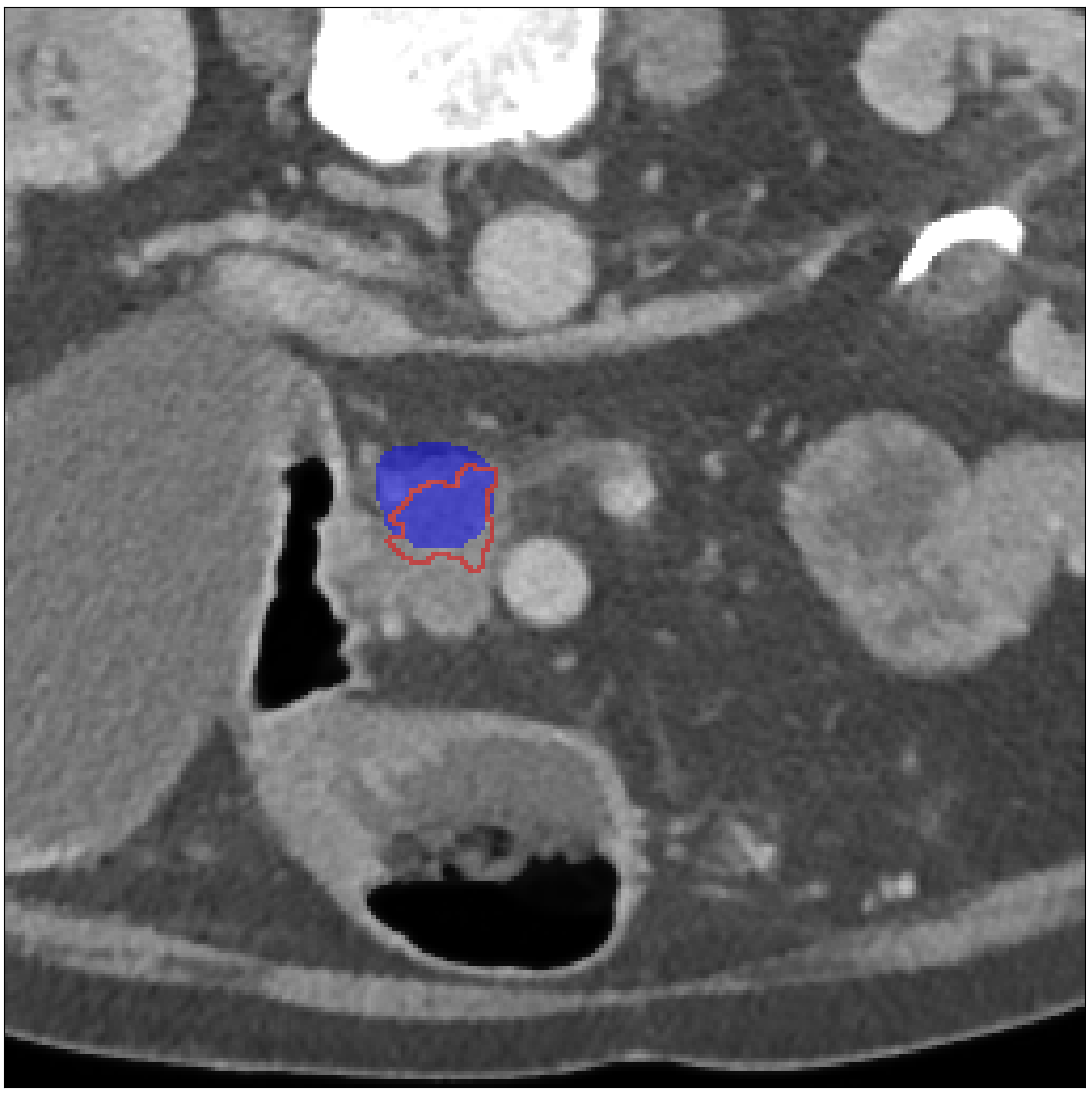}  
    \caption{\footnotesize{3D Dice: 0.44 }}
    \label{SUBFIGURE LABEL 1}
\end{subfigure}
\begin{subfigure}{.24\textwidth}
    \centering
    \includegraphics[angle=180, width=.95\linewidth]{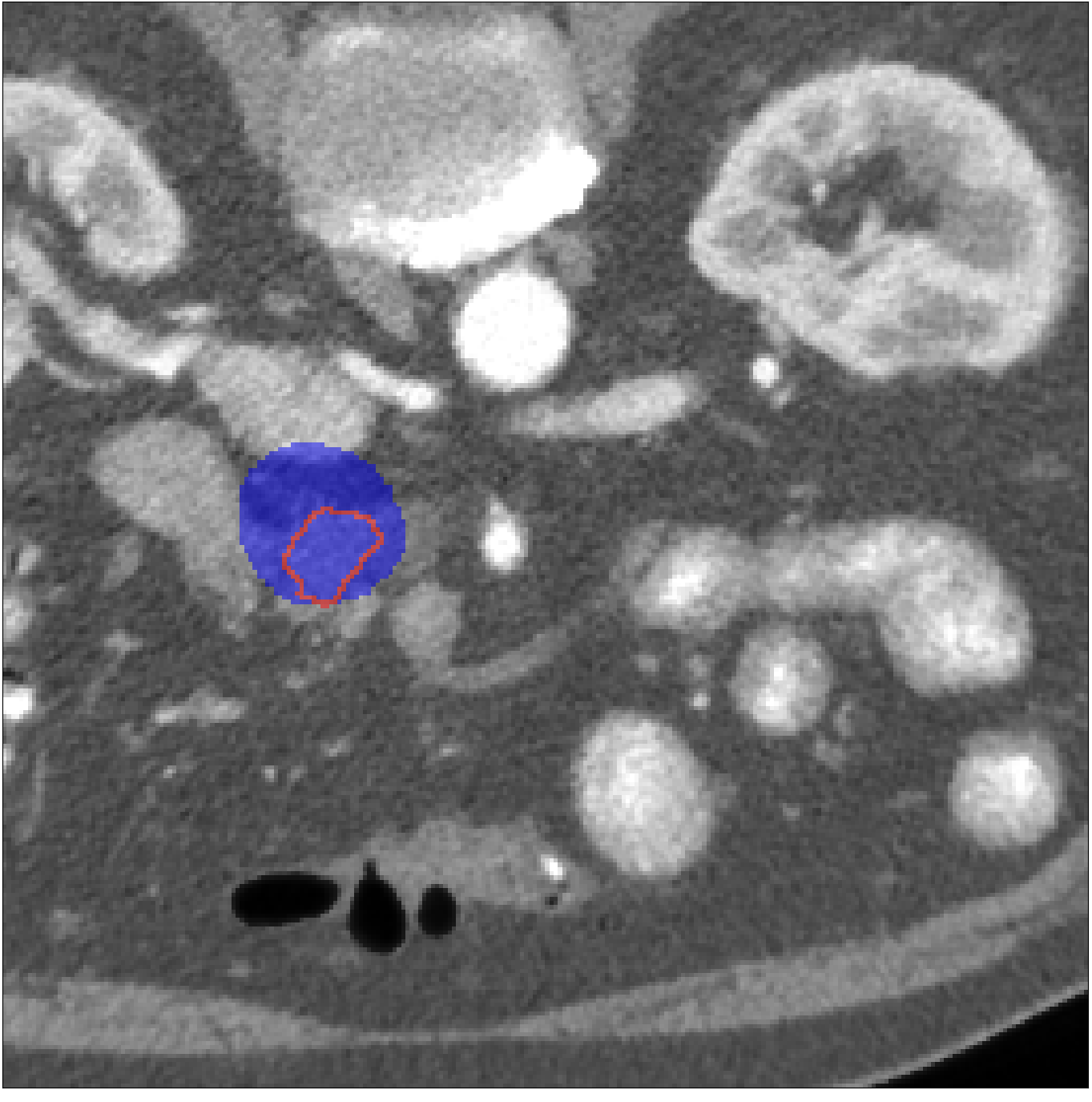}  
    \caption{\footnotesize{3D Dice: 0.29}}
    \label{SUBFIGURE LABEL 2}
\end{subfigure}
\begin{subfigure}{.48\textwidth}
    \centering
    \includegraphics[width=.95\linewidth]{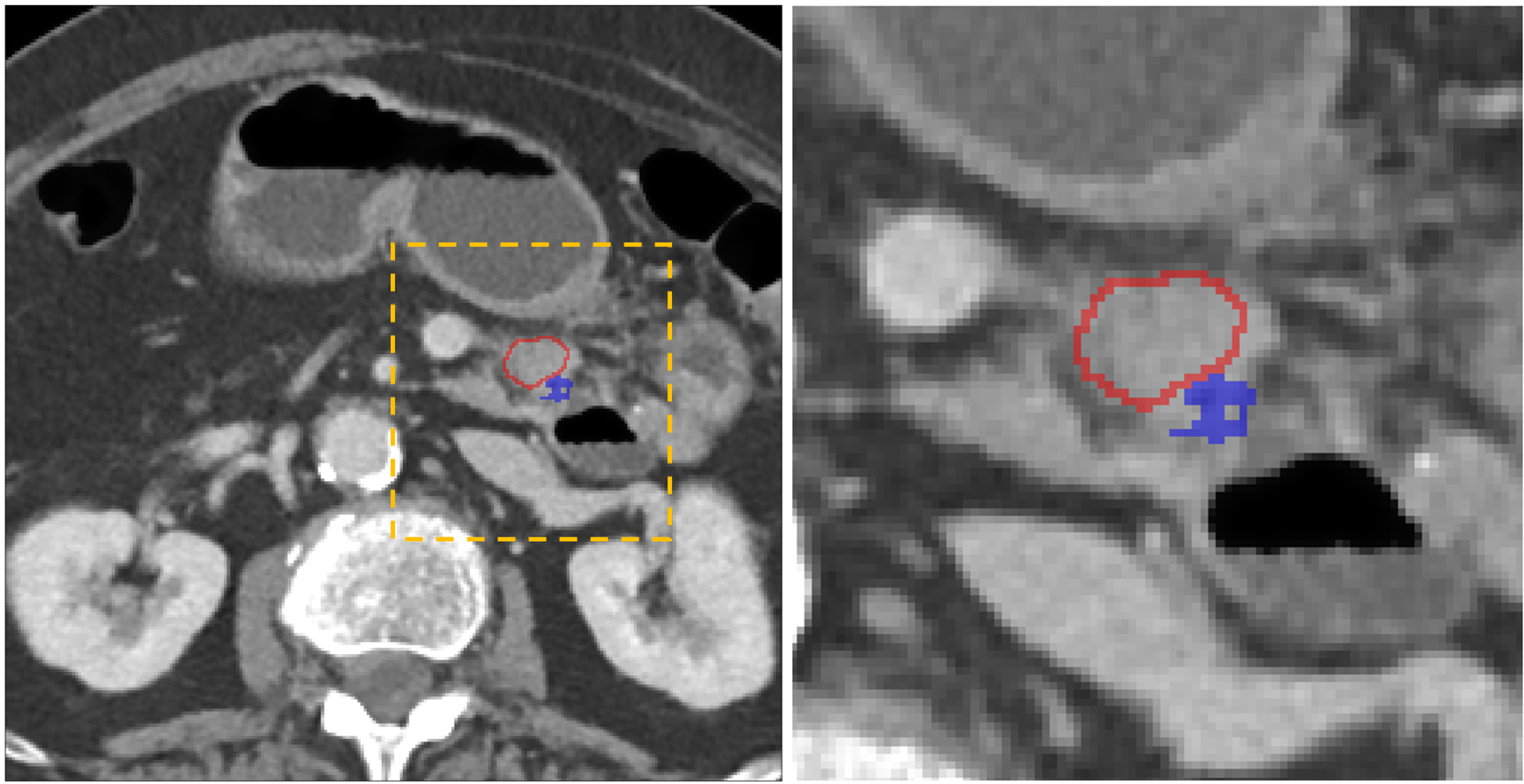}  
    \caption{\footnotesize{3D Dice: 0.01}}
    \label{SUBFIGURE LABEL 3}
\end{subfigure}}
\caption{\textit{Segmentation performance from three different cases. An example of a low-performing segmentation is visualized (\ref{SUBFIGURE LABEL 3}). The figure is best viewed in color.}\label{slice_examples} }
\label{FIGURE LABEL}}
\end{figure}

\emph{Test Set (MD) Experiments:} We observe very similar impressive performance when the model is applied to the MD dataset as test set. In two of the three folds, the models showcase 100\% sensitivity, with no tumor missed. The model from the third fold missed the tumor in one of the cases and made no prediction whatsoever (False Negative). The same tumor was predicted with a relatively high Dice score of 0.40 and 0.26 in the other two models. Averaging the sensitivity across the three models, 100\%, 100\% and 96.43\%, explains the 99$\pm$2\% sensitivity at the bottom of the table (check remark on accuracy in table caption). Example predictions on the test set can be seen in Figure~\ref{slice_examples}. Note that the aim of this study is not to achieve maximum segmentation accuracy, but rather develop a more effective, clinically-relevant and efficient method for tumor detection. Inference time using this method is 0.33\,s on a RTX~2080~Ti.\\
\indent\emph{Limitations:} The secondary features used in this work and provided as external input are acquired from the same CT scan. One would expect that a CNN would be able to extract these embedded spatial features and discover the causality between these features and the presence of the tumor. Unfortunately, this expectation is not valid. Future work should investigate these underlying causal factors and how to enable a CNN to exploit this available information.
\section{Conclusion}
Despite the eminent success of deep learning networks, even for detection of PDAC, the method presented in this work demonstrates that external tumor-indicative features can significantly boost CAD performance. We optimize a segmentation for classification and localization approach, by adding the easily obtainable and clinically valuable external secondary features used by the radiologist, to considerably improve segmentation performance. The proposed approach consists of a 3D U-Net that takes the CT scan, along with a segmentation map of the pancreas, pancreatic duct and common bile duct as input, in order to finally segment the pancreatic tumor. By integrating these indicative secondary features into the detection process, the proposed method achieves a sensitivity of 99$\pm$2\% (one cased missed), yielding 5\% gain over the previous state-of-the-art method. The proposed method also achieves a specificity of 99\% and ultimately requires no sacrifice of specificity in favor of sensitivity. In addition, the method provides further insights into the tumor location and obtain similar segmentation scores on prospectively collected and the Medical Decathlon data. Generally, this research reveals the important value of explicitly including clinical knowledge into the detection model. We suggest that future CAD methods integrate higher orders of feature information, particularly valuable clinical features, into their domain-specific problem to improve performance when such information can be identified. This method paves the way for equipping clinicians with the necessary tools to enable early PDAC detection, with the aim to ultimately improve patient care.

\clearpage
\bibliographystyle{splncs04}
\bibliography{bibliography}

\end{document}